\documentclass{article} 
\usepackage{iclr2015,times}
\usepackage{hyperref}
\usepackage{url}
\usepackage{amsmath}
\usepackage{amssymb}
\usepackage{graphicx}
\usepackage{caption}
\usepackage{subcaption}
\usepackage{epstopdf}

\title{Unsupervised Feature Learning from Temporal Data }

\author{
Ross Goroshin$^1$, Joan Bruna$^{1,2}$, Jonathan Tompson$^1$, David Eigen$^1$, Yann LeCun$^{1,2}$ \\
$^1$Courant Institute of Mathematical Science 
719 Broadway 12$^{th}$ Floor, New York, NY 10003 \\
$^2$Facebook AI Research, 770 Broadway, New York, NY 10003\\
\texttt{\{goroshin,bruna,tompson,deigen,yann\}@cims.nyu.edu}  
}

\iclrfinalcopy 

\begin{document}
\maketitle

Current state-of-the-art classification and detection algorithms rely on supervised training. In this work we study unsupervised feature learning in the context of temporally coherent video data. We focus on feature learning from unlabeled video data, using the assumption that adjacent video frames contain semantically similar information. This assumption is exploited to train a convolutional pooling auto-encoder regularized by slowness and sparsity. We establish a connection between slow feature learning to metric learning and show that the trained encoder can be used to define a more temporally and semantically coherent metric. 

Our main assumption is that data samples that are temporal neighbors are also likely to be neighbors in the latent space. For example, adjacent frames in a video sequence are more likely to be semantically similar than non-adjacent frames. This assumption naturally leads to the slowness prior on features which was introduced in SFA (\cite{SFA}). 

Temporal coherence can be exploited by assuming a prior on the features extracted from the temporal data sequence. One such prior is that the features should vary slowly with respect to time. In the discrete time setting this prior corresponds to minimizing an $L^p$ norm of the difference of feature vectors for temporally adjacent inputs. 
Consider a video sequence with $T$ frames, if $z_t$ represents the feature vector extracted from the frame at time $t$ then the slowness prior corresponds to minimizing $\sum_{t=1}^{T} \| z_t - z_{t-1} \|_p$. To avoid the degenerate solution $z_t = z_0 ~\mbox{for}~ t=1...T$, a second term is introduced which encourages data samples that are \emph{not} temporal neighbors to be separated by at least a distance of $m$-units in feature space, where $m$ is known as the margin. In the temporal setting this corresponds to minimizing $max(0,m-\|z_t - z_{t'}\|_p)$, where $|t-t'| > 1$. Together the two terms form the loss function introduced in \cite{DrLIM} as a dimension reduction and data visualization algorithm known as DrLIM. 
Assume that there is a differentiable mapping from input space to feature space which operates on \emph{individual} temporal samples. Denote this mapping by $G$ and assume it is parametrized by a set of trainable coefficients denoted by $W$. That is, $z_t = G_W(x_t)$. The per-sample loss function can be written as: 
\begin{equation} 
\label{eqn:drlimcrit}
L(x_t,x_{t'},W)=\left\{
                \begin{array}{ll}
                 \|G_W(x_t) - G_W(x_{t'})\|_p, &\text{if}~|t-t'| = 1  \\
                 \max(0,m-\|G_W(x_t) - G_W(x_{t'})\|_p) &\text{if}~|t-t'| > 1
                \end{array}
              \right.
\end{equation} 

The second contrastive term in Equation \ref{eqn:drlimcrit} only acts to avoid the degenerate solution in which $G_W$ is a constant mapping, it does not guarantee that the resulting feature space is informative with respect to the input. 
This discriminative criteria only depends on pairwise distances in the representation space which is a geometrically weak notion in high
dimensions. We propose to replace this contrastive term with a term that penalizes the reconstruction error of both data samples. 
Introducing a reconstruction terms not only prevents the constant solution but also acts to explicitly preserve information about the input. 
This is a useful property of features which are obtained using unsupervised learning; since the task to which these features will be applied 
is not known a priori, we would like to preserve as much information about the input as possible. 

What is the optimal architecture of $G_{W}$ for extracting slow features? Slow features are invariant to temporal changes by definition. In natural video and on small spatial scales these changes mainly correspond to local translations and deformations. Invariances to such changes can be achieved using appropriate pooling operators \cite{JoanScat,LeCun1998}. 
Such operators are at the heart of deep convolutional networks (ConvNets), currently the most successful supervised feature learning architectures \cite{ImageNet}. Inspired by these observations, let $G_{W_e}$ be a two stage encoder comprised of a learned, generally over-complete, linear map ($W_e$) and rectifying nonlinearity $f(\cdot)$, followed by a local pooling. Let the $N$ hidden activations, $h = f(W_ex)$, be subdivided into $K$ potentially overlapping neighborhoods denoted by $P_i$. Note that biases are absorbed by expressing the input $x$ in homogeneous coordinates. Feature $z_i$ produced by the encoder for the input at time $t$ can be expressed as $G_{W_e}^i(t) = \|h_t\|^{P_i}_p =\left(\sum_{j \in P_i} h_{tj}^{p} \right)^{\frac{1}{p}}$. Training through a local pooling operator enforces a local topology on the hidden activations, inducing units that are pooled together to learn complimentary features. In the following experiments we will use $p=2$. Although it has recently been shown that it is possible to recover the input when $W_e$ is sufficiently redundant, reconstructing from these coefficients corresponds to solving a phase recovery problem \cite{JoanPooling} which is not possible with a simple inverse mapping, such as a linear map $W_d$. Instead of reconstructing from $z$ we reconstruct from the hidden representation $h$. This is the same approach taken when training group-sparse auto-encoders \cite{groupSparsity}. In order to promote sparse activations in the case of over-complete bases we additionally add a sparsifying $L_1$ penalty on the hidden activations. Including the rectifying nonlinearity becomes critical for learning sparse inference in a hugely redundant dictionary, e.g. convolutional dictionaries \cite{LISTA}. The complete loss functional is: 
\begin{equation}
\label{eqn:loss}
L(x_t,x_{t'},W)= \sum_{\tau = \{t,t'\}} \left(\|W_d h_\tau - x_\tau\|^2 + \alpha|h_\tau| \right)+\beta \sum_{i=1}^K \left| \|h_t \|^{P_i} - \|h_{t'}\|^{P_i} \right|
\end{equation} 

\begin{figure}
  \centering
  \begin{subfigure}[b]{0.45\textwidth}
        \includegraphics[width=\textwidth]{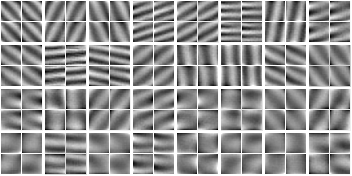}
        \caption{}
        \label{fig:pooldec}
  \end{subfigure}
  \begin{subfigure}[b]{0.45\textwidth}
        \includegraphics[width=\textwidth]{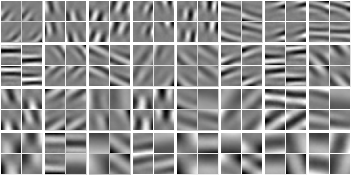}
        \caption{}
        \label{fig:pooll1dec}
  \end{subfigure}
  \caption{Pooled decoder dictionaries learned without (a) and with (b) the $L_1$ penalty using (\ref{eqn:loss}).}
  \label{fig:sfpool}
\end{figure}

\begin{figure} 
\centering
\includegraphics[scale=0.75,trim = 15 350 290 39, clip]{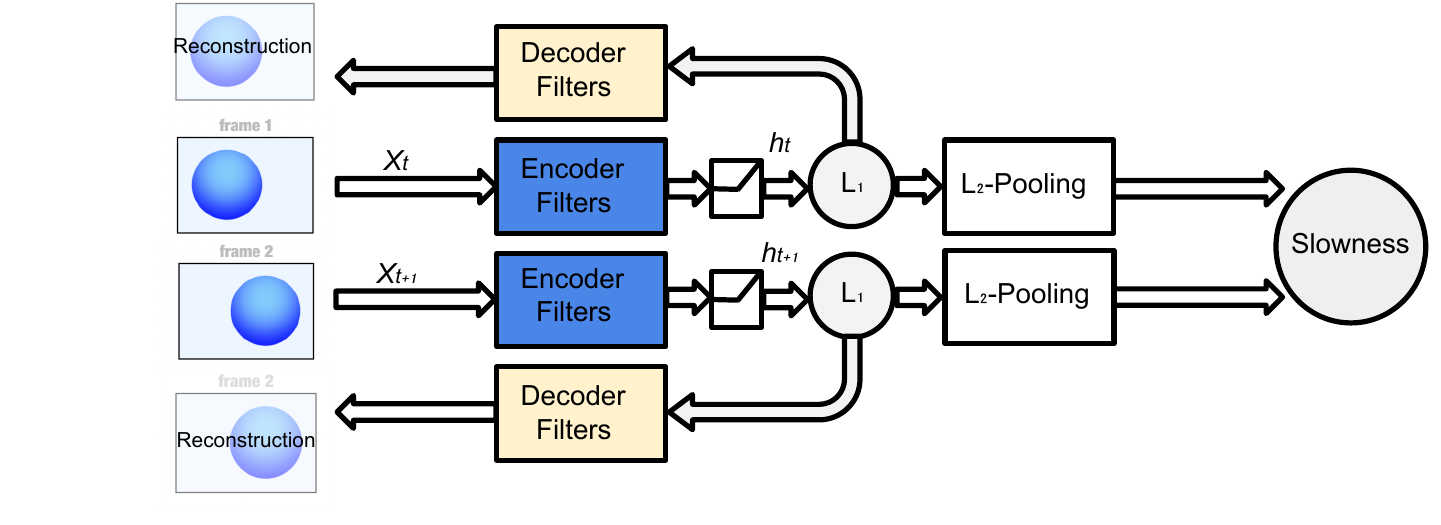}
\caption{Block diagram of the Siamese convolutional model trained on pairs of frames. \label{fig:diagram}}  
\end{figure}

Figure \ref{fig:diagram} shows a convolutional version of the proposed architecture and loss. 
By replacing all linear operators in our model with convolutional filter banks and including spatial pooling, translation invariance need not be learned \cite{LeCun1998}. In all other respects the convolutional model is conceptually identical to the fully connected model described in the previous section. 

{\small
\bibliographystyle{iclr2015}
\bibliography{iclr2015}
}

\end{document}